\documentclass{article} % For LaTeX2e
\usepackage{iclr2025_conference,times}

\usepackage{amsmath,amsfonts,bm}

\def\eqref#1{equation~\ref{#1}}
\def\1{\bm{1}}

\DeclareMathAlphabet{\mathsfit}{\encodingdefault}{\sfdefault}{m}{sl}
\SetMathAlphabet{\mathsfit}{bold}{\encodingdefault}{\sfdefault}{bx}{n}

\usepackage{hyperref}
\usepackage{url}

\usepackage{graphicx}
\usepackage{fancyhdr}
\usepackage{booktabs}
\usepackage{multirow}
\usepackage{xcolor}
\usepackage[normalem]{ulem}
\newcommand{\gain}[1]{\textcolor{green!60!black}{#1}}
\newcommand{\loss}[1]{\textcolor{red}{#1}}
\usepackage{makecell}

\title{LayerBoost: Layer-Aware Attention Reduction for Efficient LLMs}

\author{
Mohamed Ali Souibgui\thanks{Corresponding author: mohamed.souibgui@openchip.com}~~~~~Jan Fostier~~~~~Rodrigo Abadía-Heredia~~~~~Bohdan Denysenko
\vspace{-3mm}
\And
~Christian Marschke ~~~~~ Igor Peric \\ \\
~Openchip \& Software Technologies, S.L.
}
\iclrfinalcopy % Uncomment for camera-ready version, but NOT for submission.
\fancypagestyle{firstpage}{
  \fancyhf{}
  \lhead{\includegraphics[width=4cm]{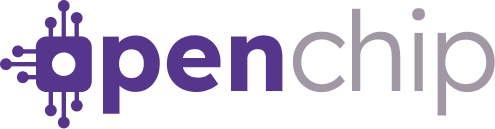}}
}
\begin{document}

% \pagestyle{fancy}
% \fancyhead[L]{}
\thispagestyle{firstpage}

\maketitle

\begin{abstract}
    Transformers are mostly relying on softmax attention, which introduces quadratic complexity with respect to sequence length and remains a major bottleneck for efficient inference. Prior work on linear or hybrid attention typically replaces softmax attention uniformly across all layers, often leading to significant performance degradation or requiring extensive retraining to recover model quality.

    This work proposes \textbf{LayerBoost}, a layer-aware attention reduction method that selectively modifies the attention mechanism based on the sensitivity of individual transformer layers. It first performs a systematic sensitivity analysis on a pretrained model to identify layers that are critical for maintaining performance. Guided by this analysis, three distinct strategies can be applied: retaining standard softmax attention in highly sensitive layers, replacing it with linear sliding window attention in moderately sensitive layers, and removing attention entirely in layers that exhibit low sensitivity. 

    To recover performance after these architectural modifications, we introduce a lightweight distillation-based healing phase requiring only 10M additional training tokens. LayerBoost reduces inference latency and improves throughput by up to 68\% at high concurrency, while maintaining competitive model quality. It matches base model performance on several benchmarks, exhibits only minor degradations on others, and significantly outperforms state-of-the-art attention linearization methods. These efficiency gains make our method particularly well-suited for high-concurrency serving and hardware-constrained deployment scenarios, where inference cost and memory footprint are critical bottlenecks.

\end{abstract}

\section{Introduction}\label{sec:intro}
Large Language Models (LLMs) are at the core of many recent advances in artificial intelligence, enabling breakthroughs in natural language processing, reasoning, coding, and multimodal understanding~\citep{achiam2023gpt,jiang2024mixtral,xiaomi2025mimo,yang2025qwen3}. Most of these modern LLMs are built upon the transformer architecture introduced by~\citet{vaswani2017attention}, whose central component is the self-attention mechanism. Self-attention is highly effective because it allows models to capture long-range dependencies by dynamically weighting interactions between all the tokens in a sequence.

Despite its effectiveness, the standard softmax attention mechanism exhibits quadratic complexity with respect to sequence length, making it a major bottleneck for efficient training and inference~\citep{vaswani2017attention}. As sequence lengths grow and deployment scenarios require higher concurrency, this quadratic scaling significantly increases memory usage and computational cost, limiting the practicality of softmax attention transformers in long-context and large-scale serving environments~\citep{sun2025speed}.

To address this limitation, several alternatives to softmax attention have been proposed with the goal of reducing its complexity. One line of work introduces linear attention mechanisms, which approximate or reformulate the softmax operation to achieve linear complexity with respect to sequence length~\citep{katharopoulos2020transformers, yang2024gated, yanggated}. Another direction replaces attention with state-space models (SSMs), which process sequences using recurrent formulations that scale efficiently to long contexts~\citep{gu2024mamba, dao2024transformers, lahoti2026mamba}.
While these approaches offer promising efficiency improvements over standard softmax attention, they typically require training models from scratch, which demands substantial computational resources and specialized training infrastructure. Furthermore, despite their efficiency advantages, these architectures can struggle on complex benchmarks that require reasoning and implicit token interactions, settings in which full softmax attention mechanisms remain highly effective. As a result, investing in training entirely new linear architectures from scratch may not always represent the most cost-effective path toward efficient large language models.

An alternative approach is attention modification, which starts from an already pretrained softmax attention based transformer model and replaces all its attention blocks with lower complexity alternatives, e.g., linear attention~\citep{chen2024dijiang,dao2024transformers}. However, such architectural modifications often introduce significant performance degradation, requiring heavy re-training with billions of tokens to restore model quality during a \emph{healing} phase.

More recently, several works have explored \emph{hybrid attention} architectures, where different attention mechanisms are combined within the same model. For instance, some layers retain standard softmax attention, while others use sliding-window attention (SWA) and/or linear attention variants~\citep{zhang2025lolcats, lan2025liger}. Although these models are not fully linear, they significantly lower the overall computational cost in practice and reduce the amount of data required during the healing phase (only millions of tokens to recover an acceptable performance). Furthermore, recent studies suggest that training such hybrid architectures on a much larger number of tokens can further improve their accuracy and reduce the performance gap with fully softmax-based models~\citep{pan2025spikingbrain}. But, despite these advances, an important limitation remains in how layers are selected for subquadratic transformation. Existing approaches typically choose which layers to modify using uniform or random selection, which can create a large performance gap between the modified architecture and the original pretrained model, making the subsequent healing process more difficult and expensive.

To address this issue, we propose LayerBoost, a layer-aware attention reduction method for constructing more efficient LLMs from pretrained softmax-based models. Rather than modifying layers arbitrarily, we first perform a sensitivity analysis to measure the impact of replacing or removing attention in each transformer layer. This analysis provides a principled basis for identifying where attention can be safely simplified. Guided by these insights, we build a hybrid architecture that retains softmax attention in highly sensitive layers, replaces it with sliding-window attention in moderately sensitive layers, and removes attention entirely in layers with minimal sensitivity. Starting from this favorable initialization, we significantly reduce the performance gap between the modified architecture and the original model. To further recover performance, we introduce a lightweight distillation-based healing phase requiring only 10M training tokens. The resulting model is substantially faster than its fully softmax-based version while maintaining competitive performance on several benchmarks. Our main contributions are:
(i) {Layer-sensitive attention linearization:} We introduce a sensitivity-driven layer selection strategy to identify transformer layers that are critical for maintaining model performance, enabling principled attention modification instead of arbitrary selection. Our approach achieves up to {68\% higher throughput} compared to standard softmax attention at high concurrency.
(ii) {Efficient recovery via lightweight distillation:} We show that only {10M tokens} of distillation-based training are sufficient to recover model performance after architectural modification, corresponding to just {$2.78 \times 10^{-5}\%$} of the original training tokens.
(iii) {Superior performance retention:} LayerBoost consistently outperforms existing attention linearization approaches across multiple benchmarks, achieving stronger overall performance retention while maintaining competitive accuracy.

\section{Related Work}\label{sec:sota}

\subsection{Efficient Attention Mechanisms}

% The quadratic complexity of the softmax attention mechanism introduced in the transformer architecture~\citep{vaswani2017attention} has motivated extensive research on improving the efficiency of sequence models. A large body of work has focused on developing alternative attention mechanisms that reduce the computational and memory complexity associated with long-context processing.

The quadratic complexity of softmax attention in transformers~\citep{vaswani2017attention} has motivated extensive work on more efficient mechanisms for long-context processing. One prominent direction is \emph{linear attention}, which reformulates the attention computation to achieve linear complexity $O(n)$ with respect to sequence length $n$. Early approaches, such as linear transformers~\citep{katharopoulos2020transformers}, approximate the softmax kernel with feature maps that allow attention to be computed using associative operations. Subsequent work further improved these methods through better kernel approximations, feature maps learning and gating mechanisms~\citep{zhang2024the,yang2024gated,yanggated}. Another class of approaches improves attention efficiency by restricting the receptive field of the attention mechanism. Instead of allowing each token to attend to the entire sequence, \emph{local or sliding-window attention} limits attention computation to a fixed neighborhood around each token. This strategy reduces the computational complexity from $O(n^2)$ to $O(nw)$, where $w$ denotes the window size. Such mechanisms have been successfully employed in models such as Longformer~\citep{beltagy2020longformer}, BigBird~\citep{zaheer2020big}, and more recently in efficient large language models that combine local and global attention patterns~\citep{team2024gemma,yuan2025native,agarwal2025gpt}. However, while SWA significantly reduces computational cost and preserves strong performance in many tasks, its restricted receptive field can limit the model's ability to capture long-range dependencies when used in isolation.  Another line of research replaces attention mechanisms entirely with alternative sequence modeling architectures. In particular, {state-space models} have recently gained attention as an efficient alternative to transformers. Models such as Mamba~\citep{gu2024mamba} and its subsequent extensions~\citep{dao2024transformers,lahoti2026mamba} process sequences using recurrent state-space dynamics that scale linearly with sequence length. These architectures demonstrate strong efficiency properties and competitive performance on several benchmarks. However, similar to linear attention models, they typically require large-scale training from scratch and substantial computational resources.
Hence, while these approaches remarkably reduce computational and spatial complexity of the attention block, they often require training models from scratch and tend to exhibit performance gaps compared to standard softmax attention models.

\subsection{Attention Reduction}

An alternative line of work aims to improve efficiency without training entirely new models. Rather than designing architectures from scratch, these approaches focus on attention reduction, replacing the attention mechanisms in pretrained transformers with more efficient variants. Early methods substituted all softmax attention layers with linear alternatives, such as linear attention~\citep{mercat2024linearizing,chen2024dijiang}, SSMs~\citep{dao2024transformers,wang2024mamba}, or SWA~\citep{yu2025swaa}. However, these wholesale replacements often lead to significant performance degradation, necessitating costly retraining on large-scale data to recover the original model capabilities. More recent works have explored \emph{hybrid attention architectures} that combine multiple attention mechanisms within the same model. In these approaches, some layers retain full softmax attention while others use more efficient variants such as sliding-window attention or linear attention~\citep{zhang2025lolcats,lan2025liger}. By using this hybrid approach and even preserving global attention in a subset of layers, these methods significantly reduce the performance gap introduced by attention modification and require considerably fewer tokens during the healing phase. Despite these advances, an important challenge remains in determining which layers should be modified. Existing approaches typically rely on random layer selection when replacing attention mechanisms~\citep{lan2025liger, yu2025swaa}. This often leads to a large performance gap between the modified architecture and the original model, making the subsequent healing process more difficult. More recently, \citet{gu2025jetnemotron} propose a framework that trains a  ``once-for-all''  supernetwork on top of a pretrained full-attention model and performs a post-hoc search to identify the optimal placement of full-attention layers. But, this approach still requires 
training a large supernetwork over billions of tokens, 
followed by task-dependent layer selection, which is computationally expensive.

\section{Methodology}\label{sec:method}

\subsection{Problem Formulation}
Consider a pretrained transformer composed of $L$ layers, each containing a self-attention module followed by an MLP. Given an input sequence
$x = (x_1, \dots, x_n)$,
standard self-attention computes:

\begin{equation}
\text{Attn}(Q,K,V) =
\text{softmax}\left(\frac{QK^\top}{\sqrt{d}}\right)V,
\end{equation}

where $Q,K,V \in \mathbb{R}^{n \times d}$ are the query, key, and value projections of the input tokens and $d$ is the hidden dimension. While effective, this mechanism introduces quadratic complexity $\mathcal{O}(n^2)$ with respect to sequence length $n$, when both forming and applying the attention matrix $QK^\top$. This becomes a major computational bottleneck for long contexts as the inference speed is highly affected.

Our goal is to transform a pretrained \emph{softmax-based} transformer into a more efficient architecture, while preserving most of its performance and requiring minimal post-training. Formally, given a pretrained model $f_{\theta}$, we aim to construct a modified model $f_{\hat{\theta}}$ by replacing attention mechanisms in selected layers with more efficient alternatives while minimizing the performance degradation:

\begin{equation}
\min_{\hat{\theta}} \; \mathcal{L}\big(f_{\hat{\theta}}(x), f_{\theta}(x)\big),
\end{equation}

where $\mathcal{L}$ denotes the discrepancy between the outputs of the modified model and the original pretrained model, and $f_{\theta}$ acts as a reference, or teacher model, during the adaptation phase.

\begin{figure}[t]
\centering
\includegraphics[width=0.95\linewidth]{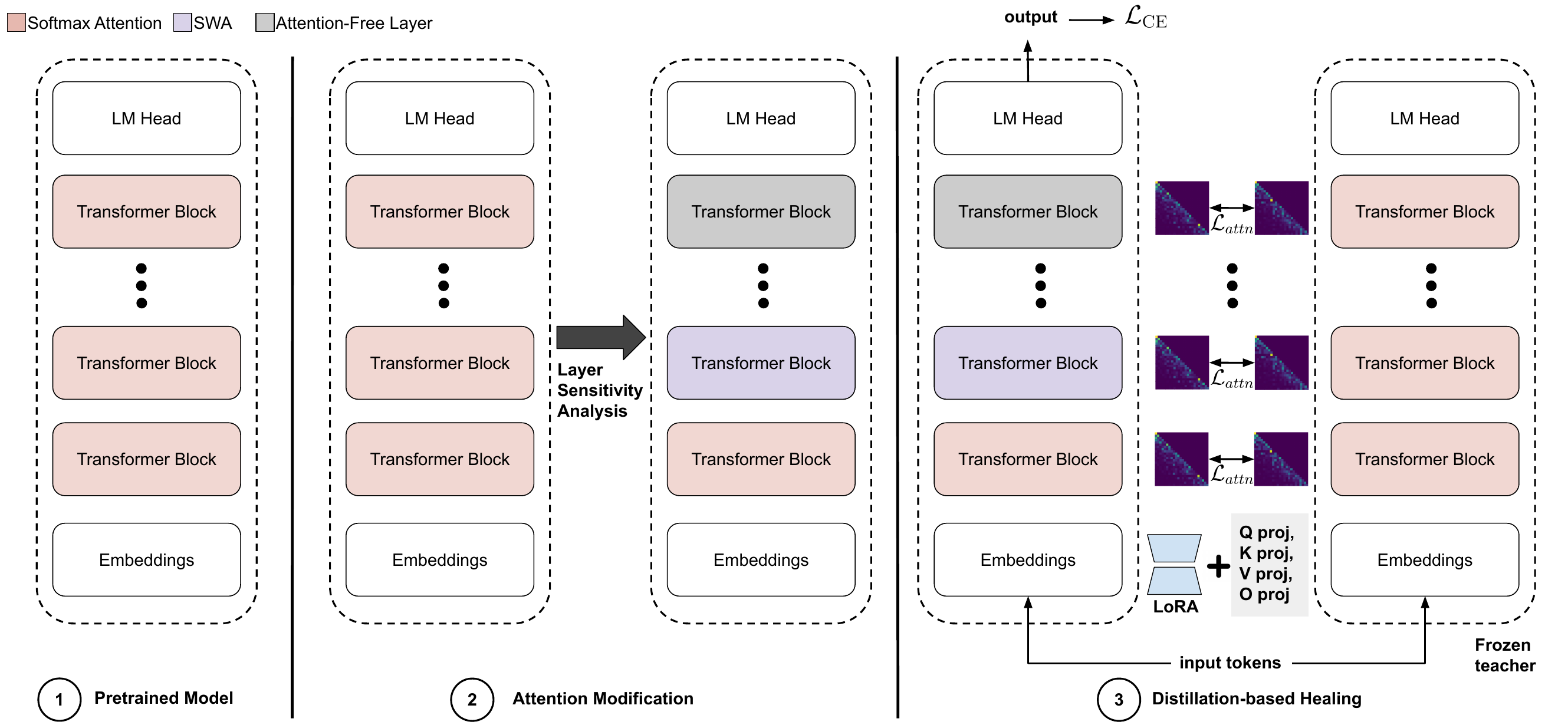}
\caption{\textbf{Overview of LayerBoost}. A pretrained transformer model is first analyzed to identify layer sensitivity to attention modifications. Based on this analysis, attention mechanisms are selectively replaced with sliding-window attention (SWA) or removed entirely in low-sensitivity layers, resulting in a hybrid architecture with subquadratic complexity. The modified model is then refined through a lightweight distillation-based healing phase using the original pretrained model as a frozen teacher. During this phase, parameter-efficient adaptation is performed using  Low-Rank Adaptation (LoRA), where only the query, key, value, and output projection matrices of the attention layers are updated, while the MLP layers and remaining parameters of the student model remain frozen.}
\label{fig:architecture}
\end{figure}

\subsection{Layer Sensitivity Analysis}

A key challenge in attention reduction is determining which layers can tolerate attention modifications without significantly affecting model performance. Instead of modifying layers arbitrarily ~\citep{lan2025liger}, we perform a layer sensitivity analysis on the pretrained model. For each transformer layer $l$, we temporarily replace the attention mechanism with an identity mapping (i.e., removing attention) and measure the resulting impact on downstream evaluation benchmarks.

Let $M$ denote the pretrained model and $M_{-l}$ the model where attention in layer $l$ is removed. The sensitivity of layer $l$ is estimated by the performance drop measured as follows,

\begin{equation} \label{eq: sensitivity}
S_l = E(M) - E(M_{-l}),
\end{equation}

where $E(\cdot)$ denotes the evaluation score on a set of commonsense reasoning benchmarks~\citep{bisk2020piqa,levesque2012winograd,clark2018think}.
The resulting sensitivity scores reveal that layers contribute unequally to model performance. Some layers are highly sensitive, where removing attention leads to a substantial performance drop, while others exhibit moderate sensitivity with only partial degradation. A third group shows low sensitivity, where attention removal has minimal impact. We use this as a prior to guide a constrained search over layer configurations.

Specifically, we impose a fixed budget on the number of layers that retain full softmax attention, prioritizing candidates from the high-sensitivity group. Using this constraint, all layers are progressively modified, and each candidate configuration is evaluated on downstream benchmarks. Modifications of layers are retained only if they do not introduce significant performance degradation; otherwise, they are reverted. This results in a performance-constrained search over feasible configurations to identify a favourable trade-off between efficiency and performance.
\subsection{Hybrid Attention Architecture}

Based on the architecture search, we construct a hybrid subquadratic model composed of: 

\noindent\textbf{Softmax Attention Layers.} Retain the original attention mechanism with $\mathcal{O}(n^2)$ complexity.

\noindent\textbf{Sliding-Window Attention Layers.}
To reduce complexity, we replace attention in moderately sensitive layers, identified with Eq.~\ref{eq: sensitivity}, with sliding-window attention (SWA), where each token attends only to a local window of size $w$:

\begin{equation}
\text{SWA}(Q,K,V)_i =
\sum_{j \in N_w(i)} \alpha_{ij} V_j,
\end{equation}

where $N_w(i)$ denotes the set of tokens within a window of size $w$ on the left of current token $i$. This reduces complexity from
$\mathcal{O}(n^2)~\text{to}~\mathcal{O}(nw)$, where typically $w \ll n$ (we use a window size of 64 tokens, same as~\cite{lan2025liger,zhang2025lolcats}).

\noindent\textbf{Attention-Free Layers.} In layers with very low sensitivity, the attention mechanism is removed and replaced by an identity mapping,  which further reduces computational cost.

% Together, these modifications produce a subquadratic transformer architecture while preserving attention where it is most needed.

\subsection{Model Healing}

The structural changes introduced by modifying the attention blocks across layers lead to performance degradation. To recover the model quality, we perform a lightweight \emph{healing phase} using knowledge distillation~\citep{44873}. Let $f_T$ denote the frozen \emph{teacher} model with full softmax attention and $f_S$ the modified \emph{student} model. The student model is trained using a combination of token-level supervision and attention-level distillation. Thus, the first component of the training objective is the standard language modeling loss:

\begin{equation}
\mathcal{L}_{CE} = \mathcal{L}_{CE}(y, f_S(x)),
\end{equation}

where $y$ denotes the ground-truth tokens. Moreover, since modifying the attention blocks alters how information is propagated across tokens, potentially disrupting the relational structure learned by the base model,
%  In particular, attention distributions encode how tokens interact and aggregate contextual information, which is a 
%  critical component in transformers. To 
we mitigate this effect by introducing an additional distillation objective that aligns the internal attention patterns of the student with those of the teacher in the most sensitive layers.
Let $A_T^{(l)}$ and $A_S^{(l)}$ denote the attention matrices of the teacher and student at layer $l$. Each row of the softmax-normalized attention matrix defines a probability distribution over input tokens, capturing how a token attends to the rest of the sequence. To preserve these interaction patterns, we minimize the row-wise Kullback-Leibler divergence between the teacher and student attention distributions:

\begin{equation}
\mathcal{L}_{attn} =
\sum_{l \in {L}_s}
\frac{1}{n}
\sum_{i=1}^{n}
KL\big(A_T^{(l)}[i,:] \parallel A_S^{(l)}[i,:]\big),
\end{equation}

where ${L}_s$ denotes the set of layers that retain softmax attention and $n$ is the sequence length. The final training objective becomes

\begin{equation} \label{eq: healing_loss}
\mathcal{L} = \mathcal{L}_{CE} + \lambda \mathcal{L}_{attn}.
\end{equation}
Where $\lambda$ is a hyperparameter that controls the trade-off between the two terms, in our experiments we use$\lambda = 0.5$, which provided the best balance between stability and performance.

Moreover, to reduce training cost, we employ Low-Rank Adaptation (LoRA)~\citep{hu2022lora}. During the healing phase, only the query, key, value, and output projection matrices of the attention modules are updated, while the MLP layers and the remaining parameters of the student model remain frozen.

\section{Experiments and Results}\label{sec:experiments}
\subsection{Experimental Setup}\label{sec:expr_setup}

\textbf{Base Model.}
Our method starts from the pretrained Qwen3-4B~\citep{yang2025qwen3} model, a transformer-based language model composed of $36$ layers with standard softmax attention.
%  This model serves as the fully quadratic reference architecture.

\textbf{Derived Architecture.}
We construct a hybrid architecture by selectively modifying the attention mechanisms of the pretrained model as described in Section \ref{sec:method}.
Specifically, we measure the layer sensitivity on a set of benchmarks, including PIQA~\citep{bisk2020piqa}, WinoGrande~\citep{levesque2012winograd}, ARC-Easy, and ARC-Challenge~\citep{clark2018think}. These benchmarks are chosen as they assess both language understanding and commonsense reasoning capabilities, providing a representative signal for evaluating the impact of attention modifications.
% The observed performance degradation in comparison with the base model is used to estimate the sensitivity of each layer and determine which layers retain full softmax attention. 
% We then construct a hybrid configuration under a fixed budget, 
We impose that 33\% of layers retain full softmax attention. Concretely, this results in the following architecture: 12 layers retain full softmax attention (high-sensitivity layers), 19 layers use sliding-window attention and 5 layers remove attention entirely (low-sensitivity layers). This configuration significantly reduces the computational complexity of the model while preserving its ability to capture dependencies in the most critical layers.  Additional details on this analysis are provided in Appendix~\ref{appdx:layer_analysis}.

\textbf{Healing Training Procedure.}
After modifying the architecture, we perform a healing phase using 40M tokens, tokenized with the Qwen tokenizer~\citep{bai2023qwen}. The training data is sampled from the Dolma-3 dataset~\citep{olmo2025olmo3} while preserving the original data type distribution. Dolma-3 provides a large-scale, diverse mixture of natural data sources, including web text, academic documents, and code, which closely reflects the distribution used during pretraining. Further details regarding this training are in Appendix~\ref{appdx:training_details}.

% \textbf{Baselines.}
% To evaluate the effectiveness of LayerBoost, we compare against a range of state-of-the-art models with parameter sizes spanning from 1.5B to 9B. 

\textbf{Primary comparison.} Our main comparison is against existing {attention reduction} approaches that modify pretrained softmax-based models and heal it with minimal training tokens. This comparison directly evaluates how effectively different methods preserve the performance of the original model after architectural modification.

\textbf{Additional baselines.} For broader context, we also compare against different classes of models trained from scratch or using a significant amount of tokens, we categorize them into: Standard transformers ($O(n^2)$), linear attention and state-space models ($O(n)$) and hybrid attention models.

\subsection{Main Results}
\subsubsection{Comparison with Attention Reduction  Methods}

We begin by evaluating LayerBoost against prior methods that transform a pretrained softmax-based transformer into a more efficient architecture through attention modification. Table~\ref{tab:linearization_relative_drop} reports the results. We consider four approaches: (i) SUPRA~\citep{mercat2024linearizing}, which replaces all softmax attention layers with linear attention using learned feature maps and performs extensive post-training (100B tokens); (ii) Mamba2-Llama~\citep{wang2024mamba}, which substitutes softmax attention blocks with state space model blocks; (iii) LoLCATs~\citep{zhang2025lolcats}, which replaces all attention layers with a hybrid combination of linear attention and sliding window attention (SWA); and (iv) Liger-GLA~\citep{lan2025liger}, which adopts another hybrid strategy by combining gated linear attention (GLA) with SWA in selected layers while keeping some other layers as fully softmax. For each method, we report performance on a suite of commonsense reasoning benchmarks \citep{Hendrycks_etal_2021_mmlu}, alongside the number of tokens used during the healing phase. To enable a fair comparison despite differences in base models, we focus on the relative change in performance with respect to each model's original softmax attention baseline (reported in parentheses).

Overall, earlier approaches such as SUPRA and Mamba2-Llama exhibit substantial performance degradation across benchmarks, despite requiring large amounts of additional training. More recent hybrid methods, namely LoLCATs and Liger-GLA, significantly reduce this gap while using orders much fewer tokens, which shows the importance of using sliding window attention. However, they still suffer from notable drops on more challenging tasks, particularly MMLU. In contrast, LayerBoost achieves the smallest overall degradation, with an average drop of only \textbf{\mbox{-1.52}}, substantially improving over the previous best result of \textbf{\mbox{-4.32}} obtained by Liger-GLA. This demonstrates that our layer-aware design more effectively preserves the capabilities of the original model after architectural modification.
The advantage is especially visible on the more challenging MMLU benchmark. While prior methods incur large performance losses on this task, LayerBoost maintains significantly stronger performance. Importantly, these gains are achieved with minimal additional training. Our method requires only 40M tokens, comparable to LoLCATs and Liger-GLA, and 
significantly fewer than SUPRA. Furthermore, as shown in Table~\ref{tab:healing_ablation} of Section~\ref{sec:ablation}, strong performance can already be recovered with as few as 10M tokens, highlighting the efficiency of our approach.

\begin{table*}[t]
\centering
\resizebox{\textwidth}{!}{
\begin{tabular}{l c c c c c c c}
\toprule
\multirow{2}{*}{\textbf{Method}} & \multirow{2}{*}{\textbf{\makecell{Healing \\ Tokens (B)}}} & \textbf{PIQA} & \textbf{ARC-e} & \textbf{ARC-c} & \textbf{Wino.} & \textbf{MMLU} & \multirow{2}{*}{\textbf{Avg.\ $\Delta$$\downarrow$}} \\
\cmidrule(lr){3-7}
&&acc$\uparrow$&acc-norm$\uparrow$&acc-norm$\uparrow$&acc$\uparrow$&acc(5-shot)$\uparrow$\\
\midrule
\multicolumn{8}{c}{\textbf{Base: Mistral-7B} (PIQA: 80.6 / ARC-e: 80.7 / ARC-c: 53.9 / Wino.: 74.3 / MMLU: 62.6)} \\
\midrule
SUPRA~\citep{mercat2024linearizing} & 100 & 80.4 {\scriptsize(\loss{-0.2})} & 75.9 {\scriptsize(\loss{-4.8})} & 45.8 {\scriptsize(\loss{-8.1})} & 70.3 {\scriptsize(\loss{-4.0})} & 34.2 {\scriptsize(\loss{-28.4})} & \loss{-9.10} \\
LoLCATs~\citep{zhang2025lolcats} & 0.04 & 79.7 {\scriptsize(\loss{-0.9})} & 78.4 {\scriptsize(\loss{-2.3})} & 47.4 {\scriptsize(\loss{-6.5})} & 71.0 {\scriptsize(\loss{-3.3})} & 23.7 {\scriptsize(\loss{-38.9})} & \loss{-10.38} \\
Liger-GLA~\citep{lan2025liger} & 0.02 & 80.1 {\scriptsize(\loss{-0.5})} & 78.7 {\scriptsize(\loss{-2.0})} & 49.3 {\scriptsize(\loss{-4.6})} & 70.1 {\scriptsize(\loss{-4.2})} & 36.3 {\scriptsize(\loss{-26.3})} & \loss{-7.52} \\

\midrule

\multicolumn{8}{c}{\textbf{Base: Llama-3-8B} (PIQA: 79.4 / ARC-e: 80.1 / ARC-c: 53.2 / Wino.: 72.9 / MMLU: 65.3)} \\
\midrule
SUPRA~\citep{mercat2024linearizing} & 20 & 78.9 {\scriptsize(\loss{-0.5})} & 75.1 {\scriptsize(\loss{-5.0})} & 46.5 {\scriptsize(\loss{-6.7})} & 65.8 {\scriptsize(\loss{-7.1})} & 40.9 {\scriptsize(\loss{-24.4})} & \loss{-8.74} \\
Mamba2-Llama~\citep{wang2024mamba} & 20 & 76.8 {\scriptsize(\loss{-2.6})} & 74.1 {\scriptsize(\loss{-6.0})} & 48.0 {\scriptsize(\loss{-5.2})} & 58.6 {\scriptsize(\loss{-14.3})} & 43.2 {\scriptsize(\loss{-22.1})} & \loss{-10.04} \\
LoLCATs~\citep{zhang2025lolcats} & 0.04 & 80.1 {\scriptsize(\gain{+0.7})} & 80.4 {\scriptsize(\gain{+0.3})} & 53.5 {\scriptsize(\gain{+0.3})} & 72.9 {\scriptsize(0.0)} & 42.1 {\scriptsize(\loss{-23.2})} & \loss{-4.38} \\
Liger-GLA~\citep{lan2025liger} & 0.02 & 80.3 {\scriptsize(\gain{+0.9})} & 81.1 {\scriptsize(\gain{+1.0})} & 52.5 {\scriptsize(\loss{-0.7})} & 72.0 {\scriptsize(\loss{-0.9})} & 43.4 {\scriptsize(\loss{-21.9})} & \loss{-4.32} \\

\midrule

\multicolumn{8}{c}{\textbf{Base: Qwen3-4B} (PIQA: 74.9 / ARC-e: 78.6 / ARC-c: 53.7 / Wino.: 65.7 / MMLU: 70.2)} \\
\midrule
\textbf{LayerBoost} & {0.04} & 
{75.7} {\scriptsize(\gain{{+0.8}})} & 
77.9 {\scriptsize(\loss{-0.7})} & 
51.0 {\scriptsize(\loss{-2.7})} & 
{67.0} {\scriptsize(\gain{{+1.3}})} & 
63.9 {\scriptsize(\loss{-6.3})} & 
\textbf{\loss{-1.52}} \\

\bottomrule
\end{tabular}
}
\caption{\textbf{Comparison with prior attention linearization approaches.} Values in parentheses indicate the performance change relative to the corresponding fully softmax base model. Improvements are shown in \textcolor{green!60!black}{green} and degradations in \textcolor{red!60!black}{red}.}
\label{tab:linearization_relative_drop}
\end{table*}

\subsubsection{Comparison with SOTA LLMs}

Next, we compare against a diverse set of LLMs to better contextualize LayerBoost performance.

\begin{table*}[t]
\centering
\resizebox{\textwidth}{!}{
\begin{tabular}{llccccccccc}
\toprule
\textbf{Type} & \textbf{Model}  &  \textbf{PIQA} & \textbf{Wino.} & \textbf{ARC-e} & \textbf{ARC-c} & \textbf{OBQA} & \textbf{TruthQA} & \textbf{HellaS.}  &\textbf{MMLU} \\
\midrule

\multirow{4}{*}{$O(n^2)$}
& Qwen3-4B   & 74.86 & 65.66 & 80.76 & 50.68 & 29.40 & 54.85 & 52.31 & 68.29\\
& Qwen3-1.7B  & 72.36 & 61.16 & 72.26 & 39.93 & 28.60 & 45.88 & 46.18 & 55.70\\
& Qwen2.5-1.5B  & 75.62 & 63.22 & 75.33 & 41.38 & 32.40 & 46.58 & 50.15 & 59.74\\
& Llama-3.2-3B-Inst.  & 75.46 & 67.56 & 74.07 & 43.60 & 28.00 & 49.78 & 52.19& 60.35 \\

\midrule

\multirow{2}{*}{$O(n)$}
& Mamba-2.7B  & 74.92 & 62.67 & 69.78 & 34.22 & 29.80 & 36.03 & 49.49 & 26.58 \\
& RWKV7-W3-2.9B  & 79.43 & 71.67 & 80.22 & 48.21 & 34.00 & 43.02 & 57.23& 53.25 \\
% & GLA-2.7B & -- & -- & -- & -- & -- & -- & -- & -- &  \\
% & Delta-Net-2.7B & -- & -- & -- & -- & -- & -- & -- & -- & \\

\midrule

\multirow{7}{*}{Hybrid}
& Qwen3.5-9B  & 79.27 & 73.32 & 81.22 & 54.77 & 32.60 & 53.69 & 58.35 & 78.75\\
& Qwen3.5-4B  & 77.52 & 70.79 & 81.39 & 51.62 & 29.00 & 48.88 & 54.37 & 74.41\\
& Qwen3.5-2B  & 72.19 & 62.98 & 70.58 & 38.31 & 26.60 & 43.42 & 46.37 & 59.46 \\
& Gemma-3-4B-it  & 62.07 & 50.74 & 44.86 & 27.55 & 16.20 & 47.85 & 42.43& 38.42 \\
& Zamba2-2.7B   & 79.27 & 73.88 & 79.92 & 48.98 & 32.20 & 45.77 & 57.57 & 56.54 \\
& Jet-Nemotron-2B  & 73.29 & 64.64 & 54.67 & 37.46 & 20.80 & 46.85 & 48.07&25.09  \\
& Jet-Nemotron-4B  & 77.04 & 68.51 & 62.21 & 39.76 & 23.80 & 47.36 & 52.38& 25.44 \\
& \textbf{LayerBoost-Qwen3-4B}  & {75.73} & {66.92} & {79.50} & {48.54} & 29.80 & 48.03 & 49.98& 64.40 \\

\bottomrule
\end{tabular}
}
\caption{\textbf{Performance comparison with multiple LLMs on commonsense reasoning benchmarks.} We compare standard quadratic Transformers ($O(n^2)$), linear/state-space models ($O(n)$), and hybrid attention architectures.}
\label{tab:commonsense_results}
\end{table*}

Table~\ref{tab:commonsense_results} summarizes results across models with different architectural paradigms, including standard transformers (Qwen3~\citep{yang2025qwen3}, Qwen2.5~\citep{qwen2025qwen25technicalreport}, and Llama3~\citep{grattafiori2024llama3herdmodels}), linear/state-space models (Mamba~\citep{gu2024mamba} and RWKV7-G~\citep{peng2025rwkv}), hybrid models (Qwen3.5~\citep{qwen35blog}, Gemma-3~\citep{gemmateam2025gemma3technicalreport}, Zamba2~\citep{glorioso2024zamba2}, Jet-Nemotron~\citep{gu2025jetnemotron}, and our method). We measure accuracy using lm-eval-harness~\citep{eval-harness}. We note that both our model and Jet-Nemotron are obtained via a post-training applied to a pretrained softmax transformer. However, the scale of post-training differs significantly: we use only 40M tokens, whereas Jet-Nemotron relies on 50B tokens. In contrast, the remaining models are trained from scratch using much more data, often reaching trillions of tokens~\citep{yang2025qwen3}. As it can be seen, LayerBoost achieves competitive performance with models of similar scale. Notably, it closely matches the performance of its base model (Qwen3-4B), indicating strong preservation of pretrained capabilities, while outperforming several architectures.
% Overall, these results highlight the effectiveness of our method: it enables the construction of hybrid attention models with minimal additional healing, substantially reducing the training cost compared to both starting from scratch and prior post-training methods, while maintaining strong performance.

\subsubsection{Serving Efficiency and Scalability}
We now evaluate the efficiency gains introduced by LayerBoost in a production-like environment. We compare against the base model (Qwen3-4B), as well as Qwen3.5-4B, a recent hybrid model that combines gated softmax attention~\citep{qiugated} with linear attention~\citep{yanggated}.
%  in a realistic serving setting. In contrast to previous sections that focus on model quality, this evaluation measures end-to-end inference performance
 We deploy all models using vLLM~\citep{kwon2023efficient}, a high-performance inference engine optimized for efficient batching and KV-cache management, and generate requests using EvalScope~\citep{evalscope_2024} to simulate concurrent client workloads. This setup enables us to measure key serving metrics, including time-to-first-token (TTFT), token throughput (TPS), and request throughput (Req/s). These experiments are done on a single NVIDIA A10 GPU with 24 GB of memory. The results are reported in Table~\ref{tab:serving_efficiency}, where we vary both the number of requests and the concurrency level to simulate increasing system load. As the load increases, our model consistently outperforms the other models across all metrics. In particular, we achieve lower TTFT and higher throughput (both TPS and Req/s) across all concurrency levels. The improvements become more significant at higher concurrency (e.g., $200$), where the complexity of the used attention within the model becomes a bottleneck.

\begin{table}[t]
\centering
\resizebox{\textwidth}{!}{
\begin{tabular}{c c l  c c c}
\toprule
\textbf{\# Requests} &  \textbf{Concurrency} & \textbf{Model} & \textbf{Avg. TTFT (s) $\downarrow$} & \textbf{TPS $\uparrow$} & \textbf{Req/s $\uparrow$} \\
\midrule

\multirow{3}{*}{5}& \multirow{3}{*}{1}   &
Qwen3-4B
& 0.154 & 51.71 & \textbf{0.05} \\

& &
Qwen3.5-4B
& 10.167 & 33.47 & 0.03 \\
 
&   &
LayerBoost-Qwen3-4B
& \textbf{0.146} (\gain{-5\%}) & \textbf{54.95} (\gain{+6\%}) & \textbf{0.05} (+0\%) \\

\midrule

\multirow{3}{*}{50}& \multirow{3}{*}{10}   &
Qwen3-4B
 & 0.646 & 374.33 & 0.37 \\

& &
Qwen3.5-4B
& 0.921 & 388.26 & 0.38 \\
 
&   &
LayerBoost-Qwen3-4B
& \textbf{0.593} (\gain{-8\%}) & \textbf{443.04} (\gain{+18\%}) & \textbf{0.43} (\gain{+16\%}) \\

\midrule

\multirow{3}{*}{250}& \multirow{3}{*}{50}   &
Qwen3-4B
& 3.093 & 830.55 & 0.81 \\

& &
Qwen3.5-4B
& 1.927 & 1007.68 & 0.98 \\
 
&   &
LayerBoost-Qwen3-4B
 & \textbf{1.355} (\gain{-56\%}) & \textbf{1246.78} (\gain{+50\%}) & \textbf{1.22} (\gain{+51\%}) \\

 \midrule

\multirow{3}{*}{500}& \multirow{3}{*}{100}   &
Qwen3-4B
& 42.317 & 825.17 & 0.81 \\

& &
Qwen3.5-4B
& 8.185 & 1098.15 & 1.07 \\
 
&   &
LayerBoost-Qwen3-4B
& \textbf{7.779} (\gain{-82\%}) & \textbf{1394.21} (\gain{+69\%}) & \textbf{1.36} (\gain{+68\%}) \\

 \midrule

\multirow{3}{*}{1000}& \multirow{3}{*}{200}   &
Qwen3-4B
& 151.691 & 824.48 & 0.81 \\

& &
Qwen3.5-4B
& 93.022 & 1071.62 & 1.05 \\
 
&   &
LayerBoost-Qwen3-4B
& \textbf{63.894} (\gain{-58\%}) & \textbf{1387.26} (\gain{+68\%}) & \textbf{1.35} (\gain{+67\%}) \\

\bottomrule
\end{tabular}
}
\caption{\textbf{Serving efficiency under varying concurrency levels using vLLM.} Relative improvements (in parentheses) are computed with respect to the base model Qwen3-4B. Input/output token lengths are fixed to 1024 tokens across all experiments.}
\label{tab:serving_efficiency}
\end{table}

\subsubsection{Decoding Efficiency and Memory Scaling}

\begin{figure}[h]
\centering
\includegraphics[width=\linewidth]{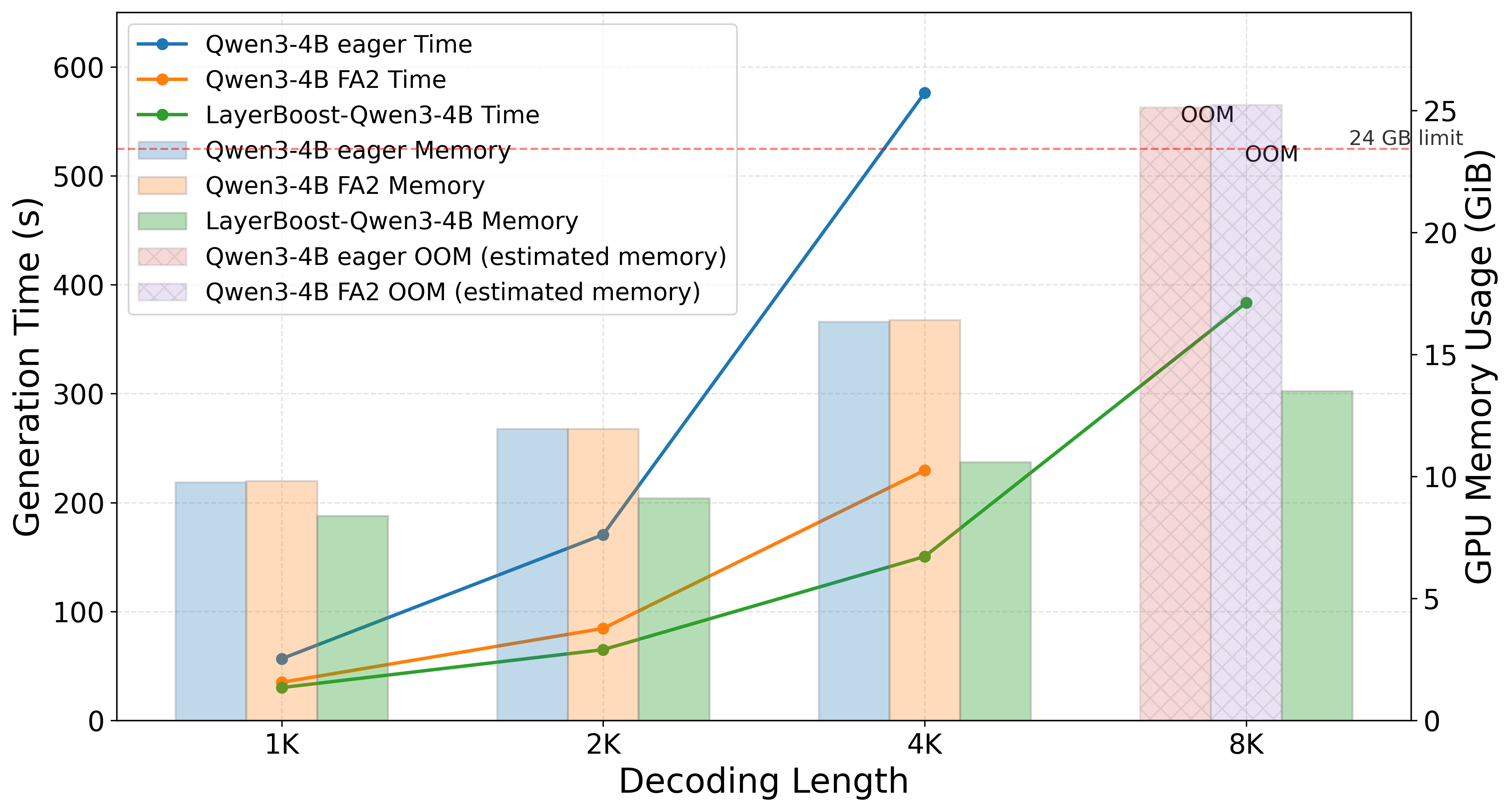}
\caption{\textbf{Decoding latency and GPU memory usage}. We use a fixed batch size of 16 and vary the decoding length.}
\label{fig:speed_mem}
% \vspace{-10pt}
\end{figure}

To further analyze the efficiency, we evaluate decoding latency and GPU memory usage as a function of sequence length. Figure~\ref{fig:speed_mem} reports generation time and peak memory  consumption for 
the base model Qwen3-4B (Huggingface eager and FlashAttention-2~\citep{dao2023flashattention} implementations) and our reduced model across increasing decoding lengths. As depicted, our method consistently achieves lower generation time compared to both baselines. While all methods exhibit increased latency as the sequence length grows, the gap becomes significantly larger at longer contexts.
More importantly, memory usage reveals a critical advantage of our approach. The baseline models exhibit rapidly increasing memory consumption due to the larger KV-cache footprint and attention overhead as sequence length increases.  Thus, both eager and FlashAttention-2 implementations exceed the 24GB GPU memory limit at 8K tokens, resulting in out-of-memory (OOM) conditions. In contrast, our model maintains a significantly lower memory footprint and remains within hardware constraints at 8K tokens.

% \begin{wrapfigure}{r}{0.60\linewidth}
% \vspace{-10pt}
% \end{wrapfigure}

% \begin{figure}[t]
% \centering
% \includegraphics[width=0.7\linewidth]{imgs/mem_and_speed.png}
% \caption{\textbf{Decoding latency and GPU memory usage of several models with different attention types}. We use a fixed batch size of 16 and vary the decoding length on a single A10 GPU with 24GB of memory.}
% \label{fig:speed_mem}
% \end{figure}

\subsubsection{Ablation Study}\label{sec:ablation}

We conduct an ablation study to evaluate the impact of the key components of our method, namely attention-level distillation and layer sensitivity-based selection, as well as the effect of the number of healing tokens. The results are in Table~\ref{tab:healing_ablation}.
First, we observe that \textbf{layer sensitivity-based selection is critical for performance}. Removing this component (``w/o Layer. Select.'') leads to a significant degradation across 4 out of the 5 benchmarks using all training budgets, with particularly large drops on ARC-Challenge and MMLU.
%  This confirms that selecting which layers to reduce based on sensitivity provides a substantially better initialization than uniform or heuristic layer replacement.
Second, \textbf{attention-level distillation further improves performance}, especially on more challenging benchmarks such as MMLU. While models trained without distillation achieve competitive results on simpler tasks, they mostly underperform when compared to our full method.
% , highlighting the importance of aligning attention distributions during the healing phase.
Finally, we analyze the effect of training tokens. Strong performance is already achieved with as few as \textbf{10M tokens}, and further improvements are obtained 
with increased training budgets. 
% Notably, our full method consistently provides the best overall performance.
% , demonstrating both the effectiveness of our design choices and the efficiency of the proposed healing procedure.

\begin{table}[t]
\centering
\small
\begin{tabular}{c l c c c c c}
\toprule
\textbf{Tokens (M)} & \textbf{Model} & \textbf{PIQA} & \textbf{Wino.} & \textbf{ARC-e} & \textbf{ARC-c} & \textbf{MMLU (0-shot)} \\
\midrule

\multirow{3}{*}{10}
& LayerBoost   & 74.76 & \textbf{67.01} & \textbf{78.58} & \textbf{47.78} & \textbf{61.91} \\
& w/o Attn. Distill. & \textbf{75.95} & \textbf{67.01} & 78.32 & 47.70 & 60.23 \\
& w/o Layer. Select. & 75.30 & 66.30 & 75.55 & 42.32 & 56.69 \\

\midrule

\multirow{3}{*}{20}
 & LayerBoost   & {75.19} & \textbf{66.61} & \textbf{79.08} & \textbf{46.93} & \textbf{62.80} \\
 & w/o Attn. Distill. & 75.19 & 65.75 & 78.62 & 46.67 & 61.79 \\
 & w/o Layer. Select. & \textbf{75.73} & 65.51 & 76.60 & 44.71 & 59.17 \\

\midrule

\multirow{2}{*}{40}
& LayerBoost   & 75.35 & 66.93 & \textbf{79.50} & \textbf{48.29} & \textbf{64.22} \\
& w/o Attn. Distill. & 75.35 &66.69&	79.29&	\textbf{48.29}	&62.28 \\
& w/o Layer. Select. & \textbf{75.57} & \textbf{67.01} & 78.03 & 45.05 & 60.13 \\

\midrule

\multirow{2}{*}{70}
& Ours   & 75.24 & \textbf{67.48} & 79.71 & 47.44 & \textbf{65.00} \\
& w/o Attn. Distill. & 75.19  & 67.40 &	\textbf{80.26} &	\textbf{49.15} &	62.99 \\
& w/o Layer. Select. & \textbf{75.46} & 66.69 & 77.86 & 46.76 & 61.31 \\

\bottomrule
\end{tabular}
\caption{\textbf{Ablation study.} We experiment the effect of training tokens and design choice during the healing phase (attention distillation and layer sensitivity analysis).}
\label{tab:healing_ablation}
\end{table}

\section{Limitations}\label{sec:limitations}

While our approach shows strong efficiency gains and competitive performance, several limitations remain. First, the results are based on specific experimental settings (model, benchmarks, hardware, serving configuration, and vLLM version) and should be considered indicative rather than general. Second, the layer sensitivity analysis relies on benchmark-driven evaluation, which may not fully capture all behaviors, especially for out-of-distribution tasks (e.g., long-form reasoning or multilingual settings). Third, the resulting hybrid architecture is model-specific, and optimal layer configurations may not transfer across models' architectures or families without re-analysis. 
Finally, our method uses a heuristic, empirical strategy to estimate layer importance, avoiding the costly combinatorial search but without guaranteeing global optimality. Nonetheless, it 
demonstrates that an effective layer 
selection can be achieved efficiently, motivating future work on more systematic search methods.

\section{Conclusion}\label{sec:conclusion}

In this paper, we introduced LayerBoost, a method for converting pretrained transformers into more efficient architectures with subquadratic complexity. Our approach improves inference efficiency, achieving higher throughput (\mbox{+68\%}) and lower TTFT (\mbox{-58\%}) while maintaining strong performance across benchmarks. Notably, it outperforms existing attention linearization methods, with only a \mbox{-1.5} average performance drop and requiring just tens of millions of tokens for recovery.

Beyond efficiency, our results suggest an alternative to training new linear or hybrid architectures from scratch. Instead, adapting pretrained softmax-based models through targeted modifications offers a more cost-effective and sustainable path to efficient LLMs.
Future work includes scaling to larger models, extending to longer contexts, and exploring improved sensitivity estimation methods.

\subsubsection*{Acknowledgments}
This work has been supported by a grant from the Program DARE\_SGA\_1 co-financed by CDTI and the Horizon Europe Research and
Innovation Framework Programme of the European Union.

\bibliography{main}
\bibliographystyle{iclr2025_conference}

\newpage
\appendix
\begin{center}
    {\LARGE {Appendix}}\\
\end{center}
\section{Layer Sensitivity Analysis}\label{appdx:layer_analysis}

In this section, we provide additional details on the layer sensitivity analysis used to guide the design of our hybrid attention architecture.

\subsection{Sensitivity Estimation Protocol}

To quantify the importance of each transformer layer, we evaluate the impact of removing the attention mechanism from individual layers of the pretrained model. Specifically, for each layer $l$, we replace the attention module with an identity mapping while keeping all other components unchanged.

We then evaluate the modified model on a set of commonsense reasoning benchmarks, including PIQA, WinoGrande, ARC-Easy, and ARC-Challenge. Let ${E}(\cdot)$ denote the average performance across these tasks. The sensitivity of layer $l$ is defined as:

\begin{equation}
S_l = {E}(M) - {E}(M_{-l})
\end{equation}

where $M$ is the original pretrained model and $M_{-l}$ is the model with attention removed at layer $l$. A higher value of $S_l$ indicates that the layer is more critical for maintaining model performance.

\subsection{Sensitivity Results}

\begin{figure}[h]
\centering
\includegraphics[width=0.85\linewidth]{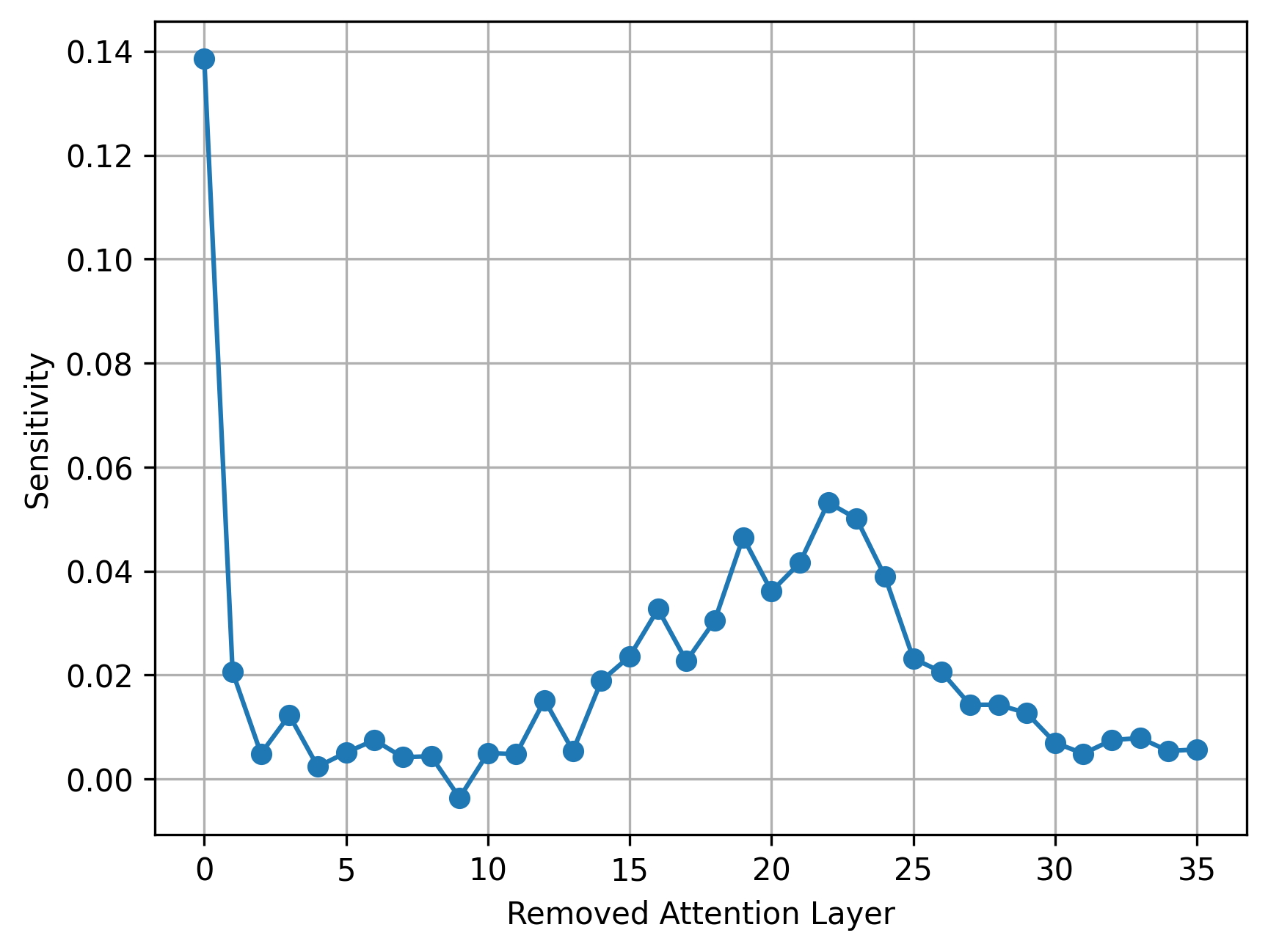}
\caption{\textbf{Performance when removing attention from individual layers}. Sensitivity denotes the average performance drop across benchmarks relative to the original model (Eq~\ref{eq: sensitivity}).}
\label{fig:layer_sensitivity}
\end{figure}

Figure~\ref{fig:layer_sensitivity} illustrates the initial sensitivity analysis of the pretrained model. It shows the performance drop of the modified model when removing attention from each layer individually  relative to the original model.
We observe that layer sensitivity is highly non-uniform across the network. Early layers and certain intermediate layers exhibit substantial performance degradation when attention is removed (e.g., layer 0), whereas others have minimal impact. These results suggest that only a subset of layers is critical for maintaining model performance.

Building on this analysis, we construct hybrid architectures under a fixed budget on the number of layers that retain full softmax attention. Specifically, we constrain a fraction of layers (33\% in our experiments) to preserve standard attention, while the remaining layers are candidates for replacement with more efficient alternatives or removal.

To identify effective configurations under this constraint, we perform a search over layer assignments guided by the sensitivity scores. Starting from the most sensitive layers, we iteratively select layers to retain full attention, while progressively modifying less sensitive layers and evaluating the resulting model performance. This procedure enables us to approximate a configuration that maximizes efficiency while minimizing performance degradation.

The resulting model is as follows:

\begin{itemize}
    \item Softmax attention layers: [0, 8, 9, 16, 17, 18, 19, 20, 21, 22, 23, 24] 
    \item SWA layers: [11, 31, 30,  3, 5, 6,10, 12, 13, 14, 27, 28, 29, 32, 33, 1, 26, 25, 15]
    \item Pruned layers: [4, 7, 34, 35, 2]
\end{itemize}

Next, in Table~\ref{tab:init_comparison}, we compare the initial performance of different linearization strategies prior to the healing phase. Existing approaches~\citep{lan2025liger} typically rely on uniform layer replacement, which leads to substantial performance degradation across benchmarks. In contrast, our sensitivity-guided, budget-constrained search yields a significantly stronger initialization that remains much closer to the base model performance.

\begin{table}[h]
\centering
\small
\begin{tabular}{lcccccc}
\toprule
\textbf{Model} & \textbf{PIQA} & \textbf{Wino.} & \textbf{ARC-e} & \textbf{ARC-c} & \textbf{Avg.}& \textbf{Loss} \\
\midrule
Qwen3-4B (Base) & 74.92& 	65.67& 	80.77& 	50.51& 	67.97	& -- \\

Uniform Linearization (Init) & 66.38 & 62.59 & 56.23 & 33.70 & 54.72 & 13.24 \\

\midrule
LayerBoost (Init) & 71.71	&63.61	&76.98	&46.50	&64.70&	3.27 \\

\bottomrule
\end{tabular}
\caption{Comparison of the initial performance of different linearization strategies before the healing phase. We compare a uniform layer replacement strategy with our layer-aware initialization. Our method starts from a stronger initialization, reducing the performance gap to the original model prior to fine-tuning.}
\label{tab:init_comparison}
\end{table}

\section{Training Details}\label{appdx:training_details}

We perform the healing phase using a pretrained Qwen3-4B model as base model, following the layer-aware attention modification described in Section~\ref{sec:method}. The model is trained on a subset of the Dolma-3 mixture dataset, using sequences of length $1024$.

Training is conducted for a single epoch with an effective batch size obtained via gradient accumulation. We use the AdamW optimizer with a learning rate of 
$1 \times 10^{-4}$, gradient clipping with a maximum norm of $1.0$, and mixed precision training.

To reduce computational cost, we adopt parameter-efficient fine-tuning using LoRA, where only the query, key, value, and output projection matrices of the attention modules are updated. We use a LoRA rank of $32$, scaling factor  $\alpha = 32$, and dropout of $0.05$. All other model parameters, including MLP layers, remain frozen.

During training, we employ a distillation-based objective (Eq.~\ref{eq: healing_loss}), combining the standard language modeling loss with attention-level distillation from the original pretrained model with a distillation weight ($\lambda = 0.5$). Checkpoints are saved at different training budgets (10M, 20M, 40M, and 70M tokens) to evaluate the effect of training data on recovery performance.

\begin{table}[h]
\centering
\small
\begin{tabular}{ll}
\toprule
\textbf{Parameter} & \textbf{Value} \\
\midrule
Base Model & Qwen3-4B \\
Training Dataset & Dolma-3 (subset) \\
Sequence Length & 1024 \\
Batch Size & 2 \\
Gradient Accumulation & 8 \\
Effective Batch Size & 16 \\
Learning Rate (Peak) & $1 \times 10^{-4}$ \\
LR Scheduler & Cosine with warmup \\
Warmup Steps & 500 \\
Gradient Clipping & 1.0 \\
\midrule
LoRA Rank ($r$) & 32 \\
LoRA Alpha & 32 \\
LoRA Dropout & 0.05 \\
LoRA Targets & $W_q, W_k, W_v, W_o$ \\
\midrule
Distillation & Yes \\
Distillation Weight ($\lambda$) & 0.5 \\
Training Tokens & 10M / 20M / 40M / 70M \\
\bottomrule
\end{tabular}
\caption{Training configuration for the healing phase.}
\end{table}

\section{Additional Results}\label{appdx:additional_results}
\paragraph{Efficiency/Utility trade-off.}

The obtained results within the paper demonstrate that our approach not only preserves model quality but also enables substantial efficiency gains in real-world deployment scenarios. To further illustrate that, we present in Figure~\ref{fig:trade-off} the trade-off between serving efficiency and model performance at high concurrency levels. We compare our model to its base Qwen3-4B and to Qwen3.5-4B. Each subplot reports benchmark accuracy against token throughput (TPS) for different models and concurrency settings. As shown, while Qwen3.5-4B attains slightly higher accuracy on some benchmarks, our model remains competitive while offering substantially better serving efficiency. Notably, at high concurrency, our method achieves up to $\sim$25--30\% higher throughput compared to Qwen3.5-4B, highlighting the benefits of our method.

\begin{figure}[t]
\centering
\includegraphics[width=0.95\linewidth]{imgs/qwen_comparison_effici_acc.png}
\caption{\textbf{Efficiency-performance trade-off at high concurrency}. 
% The x-axis reports benchmark accuracy, while the y-axis shows serving throughput (tokens per second, TPS). Each marker corresponds to a different concurrency level (50, 100, 200). 
Points located toward the top-right indicate more favorable trade-offs.}
\label{fig:trade-off}
\end{figure}

\end{document}